\documentclass{ifacconf}
\usepackage{graphicx}
\usepackage{natbib}

\usepackage{amsfonts}
\usepackage{amsmath}
\usepackage{subcaption}
\usepackage{booktabs}
\usepackage{array}
\usepackage{xcolor}

\begin{document}
\begin{frontmatter}
    
\title{Function Based Isolation Forest (FuBIF):\\ A Unifying Framework for Interpretable Isolation-Based Anomaly Detection}

\thanks[footnoteinfo]{This work was partially supported by: (1) Regione Veneto Project `Sistemi Umano centrici per Prodotti e pRocEssi Manufatturieri Evoluti' (SUPREME), PR Veneto FESR 2021--2027. Azione 1.1.1 Sub A. `Bando per il finanziamento di progetti di ricerca e sviluppo realizzati dalle Reti Innovative Regionali e dai Distretti Industriali' DGR n. 729 of June 26, 2024; (2) MICS (Made in Italy---Circular and Sustainable) Extended Partnership and received funding from Next-GenerationEU (Italian PNRR---M4C2, Invest 1.3---D.D. 1551.11--10--2022, PE00000004); (3) PNRR research activities of the consortium iNEST (Interconnected North-Est Innovation Ecosystem) funded by the European Union Next-GenerationEU (Piano Nazionale di Ripresa e Resilienza (PNRR)---Missione 4 Componente 2, Investimento 1.5---D.D. 1058 23/06/2022, ECS00000043).
}

\author[First]{Alessio Arcudi} 
\author[Second]{Alessandro Ferreri} 
\author[Third]{Francesco Borsatti}
\author[Third]{Gian Antonio Susto}

\address[First]{
    Brain, Mind and Computer Science,
    University of Padova, Italy
    (e-mail: alessio.arcudi@phd.unipd.it)
}
\address[Second]{
    Department of Mathematics, 
    Sapienza University of Roma, Italy 
    (e-mail: alessandro.ferreri@uniroma1.it)
}
\address[Third]{
    Department of Information Engineering,
    University of Padova, Italy
    (e-mail: francesco.borsatti.1@phd.unipd.it, gianantonio.susto@unipd.it)
}

\begin{abstract}
Anomaly Detection (AD) is evolving through algorithms capable of identifying outliers in complex datasets. 
The Isolation Forest (IF), a pivotal AD technique, exhibits adaptability limitations and biases.
This paper introduces the Function-based Isolation Forest (FuBIF), a generalization of IF that enables the use of real-valued functions for dataset branching, significantly enhancing the flexibility of evaluation tree construction. Complementing this, the FuBIF Feature Importance (FuBIFFI) algorithm extends the interpretability in IF-based approaches by providing feature importance scores across possible FuBIF models. This paper details the operational framework of FuBIF, evaluates its performance against established methods, and explores its theoretical contributions. An open-source implementation\footnotemark{} is provided to encourage further research and ensure reproducibility.
\end{abstract}


\end{frontmatter}
\footnotetext{repository: https://github.com/anonymous2532/FuBIF}

\section{Introduction} 
The Isolation Forest (IF) of \citet{liu2008isolation} is a staple of
anomaly detection, valued for its speed, frugal memory, and robust
performance across speech, finance, and industrial monitoring
tasks~\citep{ounacer2018using,wu2018application,kabir2023isolation}.
Yet IF's axis-parallel splits can bias scores whenever the relevant
signal lies off the coordinate axes, prompting hypersphere, extended,
and generalized variants that bend or rotate the decision
surface~\citep{hariri2019extended,choudhury2021hypersphere,lesouple2021generalized,Xu2022DeepIF}.
Almost all of these advances postpone \emph{interpretability}, yielding
patchy, method-specific heuristics that hamper root-cause analysis.

We close both gaps with \textbf{Function-Based Isolation Forest
(FuBIF)} and its explanation layer, \textbf{FuBIFFI}.  Each tree node
now partitions space by the lower/upper level sets of a randomly drawn
\(f\in\mathcal{F}\subset\{g:\mathbb{R}^d\!\to\!\mathbb{R}\}\) and a
random threshold \(\tau\).  Selecting \(\mathcal{F}\) recovers classic
IF (coordinate indicators), Extended IF (linear), Hypersphere IF
(radial), Deep-IF (neural), and other proposals, unifying them under a
single framework.  FuBIFFI then decomposes each anomaly score into
per-function, and hence per-feature, contributions by tracing forest
path lengths, delivering model-agnostic attributions with a clear
probabilistic meaning.

This lens clarifies prior “function-based IF” work, exposes their common principles, and through
FuBIFFI, links \emph{detection} directly to \emph{explanation}, enabling
actionable insight for practitioners and
regulators~\citep{doshi2017towards,rudin2019stop,arrieta2020explainable}.

The outline of the article:
Section~\ref{sec:related_works} reviews  the IF and its derivatives. 
Section~\ref{sec:mathematical_framework} explains the FuBIF and FuBIFFI interpretability framework. 
Section~\ref{sec:experiments} demonstrates the efficacy of FuBIF, comparing it with existing and novel IF-based models. 
Section~\ref{sec:future_works}, with conclusions in Section~\ref{sec:conclusion}.

% ---------------------------------------------------------------------------------------------------------
\section{Related Works}\label{sec:related_works}

The Isolation Forest (IF) algorithm \citep{liu2008isolation} is a widely used AD method. IF isolates anomalies near the tree root due to fewer splits, but is biased toward data distributed orthogonally to axes, reducing effectiveness in complex datasets. The Extended Isolation Forest (EIF) \citep{hariri2019extended} uses random hyperplanes instead of axis-aligned cuts, reducing orientation bias, but introducing other complexities. The Generalized Isolation Forest (GIF) \citep{lesouple2021generalized} improves hyperplane selection by balancing the cut distribution throughout the data range. EIF$^+$ \citep{arcudi2023exiffi} refines threshold selection using normal distribution sampling, reducing bias. The Hypersphere Isolation Forest (HIF) \citep{choudhury2021hypersphere} uses hyperspheres for space partitioning, better capturing certain data structures. The deep isolation forest (DIF) \citep{Xu2022DeepIF} utilizes neural networks to transform the data space before applying isolation forest techniques.

Key differences in tree construction and splitting criteria, that FuBIF aims to unify within a single framework, are summarized below.

For \textbf{IF} and \textbf{EIF}, the criteria is ${\mathbf{v}\cdot\mathbf{x} - \tau \leq 0}$, where \(\mathbf{v}\) is a random unit vector and \(\tau\) is chosen from the range of projected data; in \textbf{IF}, the vector is parallel to the features. 
A similar rule applies to \textbf{EIF$^+$}, but in this case, \(\tau\) is sampled from a normal distribution. In \textbf{HIF}, the branching function becomes \(\|\mathbf{x} - \mathbf{c}\|^2 - R^2 \leq 0\). 
For \textbf{DIF}, the function is \({(NN_\Theta(\mathbf{x}))}_i - \tau \leq 0\), where \(NN_\Theta\) is a randomly initialized neural network.

\section{Function Based Isolation Forest (FuBIF)}%
\label{sec:mathematical_framework}

\subsection{Isolation based Anomaly Detection}%
\label{sec:common-structure}

Isolation-based Anomaly Detection models assume abnormal samples are easier to isolate than normal ones. These methods use an ensemble of trees \(\mathcal{T}=\{T_1, T_2, \ldots, T_N\}\), trained on a random subsample \(Y_i\) drawn from a dataset \(X \subset \mathbb{R}^d\). Tree nodes correspond to recursive random splits of data $Y_i$. Anomalies tend to be isolated near the root, requiring fewer splits. The number of splits required to isolate a sample \(\textbf{x}\) is called its path length, denoted $h_{T_i}(\textbf{x})$. The Isolation Forest (IF) algorithm and its variants define the anomaly score $s(\mathbf{x},n)$ based on the expected path length across all trees in the forest, normalized by a factor $c(n)$, where $n$ is the subsample size and $c(n)$ is a normalizing factor:
\[
s(\mathbf{x},n) = 2^{\frac{-\mathbb{E}_{T_i \in \mathcal{T}}[h_{T_i}(\textbf{x})]}{c(n)}}.
\]

\subsection{Function-based Isolation Forest framework}%
\label{mathframe}

FuBIF is a generalized mathematical framework for IF models, so a class of algorithms constructed from:
\begin{itemize}
  \item a set \(\mathcal{F} \subset \{f : \mathbb{R}^n \to \mathbb{R}\}\) of data splitting functions,
  \item sampling distribution \(\rho\) over \(\mathcal{F}\),
  \item sampling distribution \(\mu\) over \(\mathbb{R}\) for thresholds.
\end{itemize}
  
FuBIF preserves forest-structure principles from~\ref{sec:common-structure}, modifying tree ramification. 
For a sub-sample \( Y \) of dataset \( X \subset \mathbb{R}^d \), a function \( f \) is selected from \( \mathcal{F} \) via distribution \( \rho \). 
A threshold \( \tau \) is drawn from distribution \( \mu \), re-centered by \( f \)'s sample distribution:
${\big\{f(\mathbf{x})| \mathbf{x} \in Y\big\}}$. 

The tree construction process starts with the root node function: ${F(\textbf{x})=f(\textbf{x})-m}$,
and then the set $Y$ is divided into two subsets:
\begin{equation*}
  \begin{split}
    L=\{\textbf{x}\in Y\ | \ F(\textbf{x})\leq 0\},\ \
    R=\{\textbf{x}\in Y\ | \ F(\textbf{x})> 0\}.        
  \end{split}
\end{equation*}

This process recursively creates left ($L$) and right ($R$) child-nodes until reaching maximum iterations or subset of single point. 
A FuBIF-tree forest is built from random dataset sub-samples, with anomaly scores determined by average point height. 
The centering of \(\mu\) relative to the output of \(f\) in FuBIF-tree construction indicates \(\mu\)'s dependence on \(f\) and $Y$.
Function-based algorithms can allow \(\mu\), \(\rho\), and \(\mathcal{F}\) to depend on node dataset, tree depth, and previous functions. 
However, in this paper, \(\mathcal{F}\) is fixed, and \(\rho\) is uniform over \(\mathcal{F}\) or re-centered uniform over a subset. Distribution \(\mu\) types are listed below, where $Y$ is the node subset and \(f\) is the chosen function.

\smallskip
\textbf{Uniform Distribution} over the range of \(f(x)\) values:
\[
\mathcal{U}\left(\left[\min_{\mathbf{x} \in Y} f(\mathbf{x}), \max_{\mathbf{x} \in Y} f(\mathbf{x})\right]\right)
\]

\smallskip
\textbf{Normal Distribution} mean/std of \(f(x)\) scaled by \(\eta\):
\[
\mathcal{N}\Big(mean\big(\{f(\mathbf{x}) | \mathbf{x} \in Y\}\big), 
\eta std\big(\{f(\mathbf{x}) | \mathbf{x} \in Y\}\big)\Big)
\]

Substantial differences arise from distribution choice for the domain for splits: one splits within data range, the other extends beyond the range based on $\eta$. We denote each algorithm by its triple as $(\mathcal{F}, \rho, \mu)$-IF.%

\subsection{Tested Examples}
This study explores generalizations of the FuBIF algorithm with splitting functions $\mathcal{F}$ as $n$-dimensional conic sections.\@\cite{choudhury2021hypersphere} explored this with the HIF algorithm using hypersphere level-sets. Table~\ref{tab:splitting_functions} lists splitting function $\mathcal{F}$, parameter distribution $\rho$, and threshold distribution $\mu$ (defined in~\ref{mathframe}).

\begin{table}[ht]
    \caption{FuBIF framing of different splitting functions.}\label{tab:splitting_functions}
    \resizebox{\linewidth}{!}{%
    \begin{tabular}{| l | l | l | l  |} 
     \hline
     \textbf{Algorithm} &  \textbf{Function} & \textbf{Parameters} & \textbf{Tresh.}\\  
     \hline
     \textbf{IF} & $f_i(\mathbf{x})=x_i$ & $i \sim \mathcal{U}(\{1,\cdots n\})$  & unif  \\ 
     \hline
     \textbf{EIF} & $f_{\mathbf{v}}(\mathbf{x})= \mathbf{v}\cdot \mathbf{x}$  & $\mathbf{v} \sim \mathcal{U}(S^{n-1})$ & unif  \\
     \hline
     \textbf{EIF$^+$} & $f_{\mathbf{v}}(\mathbf{x})= \mathbf{v}\cdot \mathbf{x}$ & $\mathbf{v}\sim \mathcal{U}(S^{n-1})$ & norm \\
     \hline
     \textbf{HIF} & $f_\mathbf{c}(\mathbf{x})=\|\mathbf{x}-\mathbf{c}\|^2$ & $\mathbf{c} \sim \mathcal{U}(\textbf{range}(Y))$ & unif \\ 
     \hline
      \textbf{Ellipse$^n$-IF} & $f_{\mathbf{c}_1,\mathbf{c}_2}(\mathbf{x})=\|\mathbf{x}-\mathbf{c}_1\| + \|\mathbf{x}-\mathbf{c}_2\|$ & $\mathbf{c}_1,\mathbf{c}_2\sim \mathcal{U}(\textbf{range}(Y))$ & both \\ 
     \hline
     \textbf{Hyper$^n$-IF} & $f_{\mathbf{c}_1,\mathbf{c}_2}(\mathbf{x})=\|\mathbf{x}-\mathbf{c}_1\| -\|\mathbf{x}-\mathbf{c}_2\|$& $\mathbf{c}_1,\mathbf{c}_2\sim \mathcal{U}(\textbf{range}(Y))$ & both\\
     \hline
     \textbf{Para$^n$-IF} & $f_{c,v}(\mathbf{x})=\|\mathbf{x}-\mathbf{c}\|+ \mathbf{v}\cdot \mathbf{x}$ & $\mathbf{c}\sim\mathcal{U}(\textbf{range}(Y))$, $\mathbf{v}\sim\mathcal{U}(S^{n-1})$& both\\ 
     \hline
     \textbf{Quad$(\lambda)$-IF} & $f_{A,\mathbf{v}}(\mathbf{x})= \mathbf{x}\cdot (A+A^t)\mathbf{x} + \mathbf{v}\cdot \mathbf{x}$ & $A_{ij}\sim\mathcal{N}(0,1), v_i\sim \mathcal{U}([-\lambda,\lambda])$& both \\
     \hline
     \textbf{NN-IF}& $f_{\Theta,b}(\mathbf{x})= NN_{\Theta,b}(\mathbf{x})$ & $\Theta \sim\mathcal{N}(0,1), b\sim \mathcal{U}([-1,1])$ & both\\
     \hline
    \end{tabular}}
\end{table}

Let $n$ be the ambient space dimension, $Y$ a set of points split at each node, and $\textbf{range}(Y)$ the hyper-rectangle enveloping $Y$, defined as:
\[
\Big[\min_{y\in Y} y_1,\  \max_{y\in Y} y_1\Big]\times \cdots \times \Big[\min_{y\in Y} y_n,\  \max_{y\in Y} y_n\Big].
\]

The first four algorithms are well-known, reformulated in the FuBIF framework, while the last five are novel:

\begin{itemize} 
    \item \textbf{Ellipse$^n$-IF}: $n$-dimensional ellipsoid. For $n=2$, it's an ellipse; for $n=3$, an ellipsoid of revolution. The equation $f_{c_1,c_2}(\mathbf{x})-\tau=0$ describes the locus of points such that the sum of the distances to the foci is constant. 
    \item \textbf{Hyper$^n$-IF}: $n$-dimensional hyperbola branch. For $n=2$, it's a hyperbola; for $n=3$, a sheet of a hyperboloid of revolution. The equation $f_{c_1,c_2}(\mathbf{x})-\tau=0$ describes the locus of points such that the difference (with sign) of the distances from $c_1$ and $c_2$ is constant. 
    \item \textbf{Para$^n$-IF}: $n$-dimensional paraboloid. Parabola for $n=2$; a paraboloid of revolution for $n=3$. The equation $f_{c,v}(\mathbf{x})-\tau=0$ describes the locus of points such that the distance from focus $c$ equals the distance from the hyperplane $\mathbf{v}\cdot \mathbf{x}-\tau=0$. 
    \item \textbf{Quad$(\lambda)$-IF}: A quadric in $n$ dimensions described by $f_{A,\mathbf{v}}(\mathbf{x}) = \mathbf{x}\cdot (A + A^t)\mathbf{x} + \mathbf{v} - \tau = 0$. The matrix $A$ is drawn from $\mathcal{N}(0,1)$, and $\mathbf{v}$'s entries are drawn from $\mathcal{U}([-\lambda, \lambda])$. 
    \item \textbf{NN-IF}: A random neural network function with scalar output. Given random weights $\Theta \sim \mathcal{N}(0,1)$ and bias $\mathbf{b} \sim \mathcal{U}(-1,1)$, the zero locus is defined by $NN_{\Theta,b}(\mathbf{x}) - \tau = 0$. 
\end{itemize}
\subsection{Motivation}
The main hypothesis is that normal data can be locally modeled as level sets of a function $f(\textbf{x}) \in \mathcal{F}$.
 
\begin{figure}[htbp] 
    \centering % <-- added
    \begin{subfigure}{0.33\linewidth}
      \includegraphics[trim={1.5cm 0.7cm 1.5cm 1.2cm}, clip, width=\linewidth]{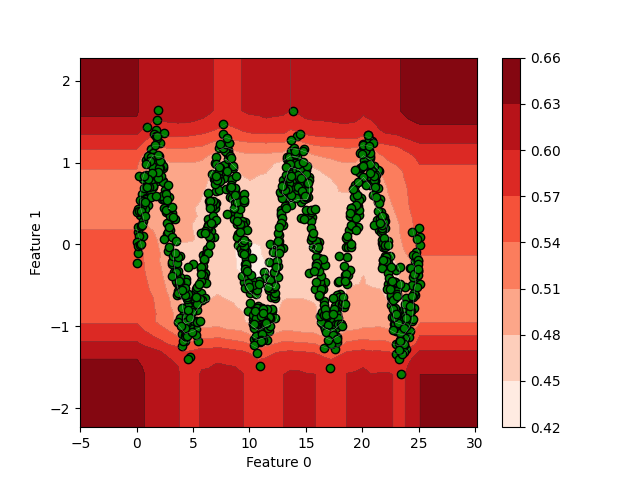}
      \caption{IF}
      \label{fig:hyper-EIF}
    \end{subfigure}\hfil % <-- added
    \begin{subfigure}{0.33\linewidth}
      \includegraphics[trim={1.5cm 0.7cm 1.5cm 1.2cm}, clip, width=\linewidth]{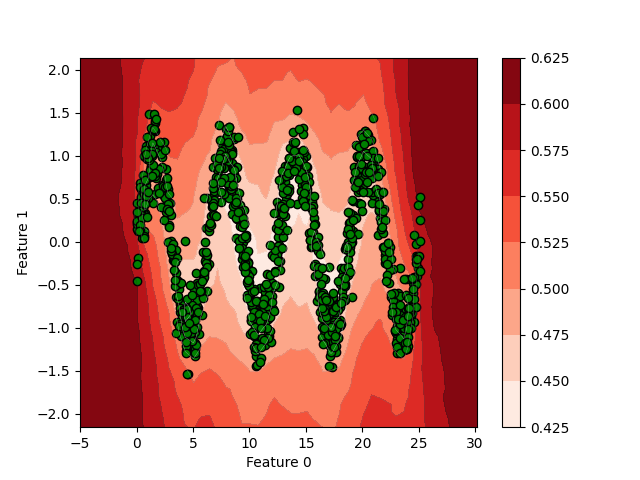}
      \caption{EIF}
      \label{fig:hyper-EIFplus}
    \end{subfigure}\hfil % <-- added
    \begin{subfigure}{0.33\linewidth}
      \includegraphics[trim={1.5cm 0.7cm 1.5cm 1.2cm}, clip, width=\linewidth]{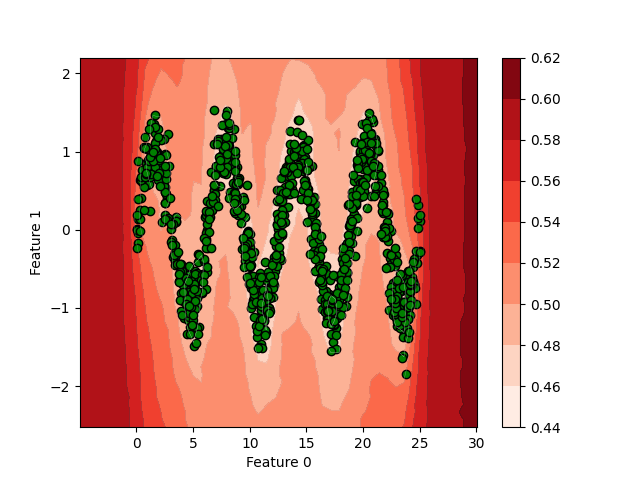}
      \caption{HIF} % hypersphere isolation forest
      \label{fig:hyper-EIF}
    \end{subfigure}\hfil % <-- added
    \begin{subfigure}{0.33\linewidth}
      \includegraphics[trim={1.5cm 0.7cm 1.5cm 1.2cm}, clip, width=\linewidth]{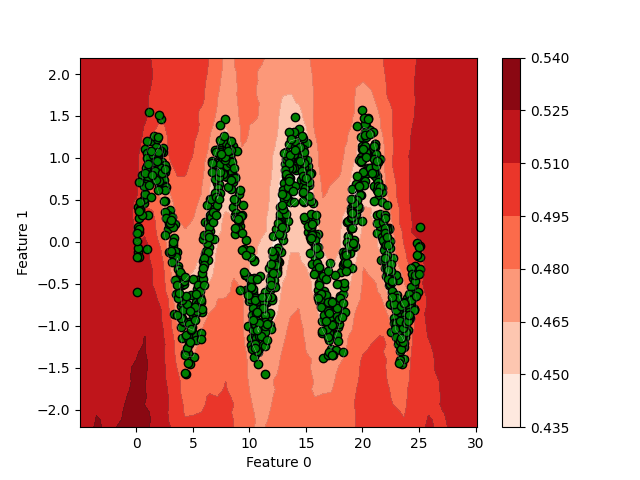}
      \caption{Hyperbola$^n$-IF}
      \label{fig:hyper-EIF}
    \end{subfigure}\hfil % <-- added
    \begin{subfigure}{0.33\linewidth}
      \includegraphics[trim={1.5cm 0.7cm 1.5cm 1.2cm}, clip, width=\linewidth]{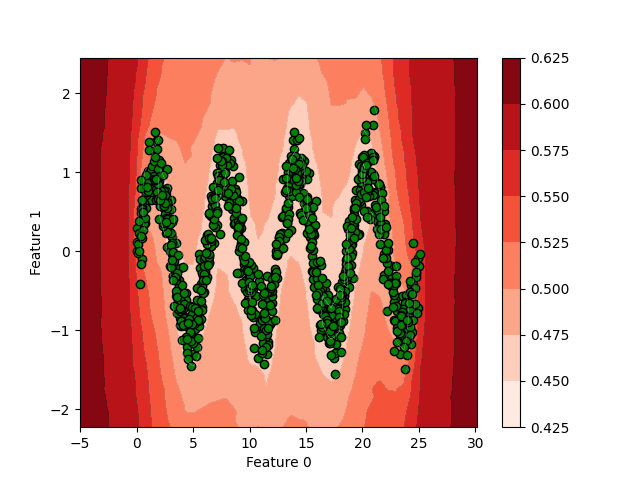}
      \caption{Ellipse$^n$-IF}
      \label{fig:hyper-EIF}
    \end{subfigure}\hfil % <-- added
    \begin{subfigure}{0.33\linewidth}
      \includegraphics[trim={1.5cm 0.7cm 1.5cm 1.2cm}, clip, width=\linewidth]{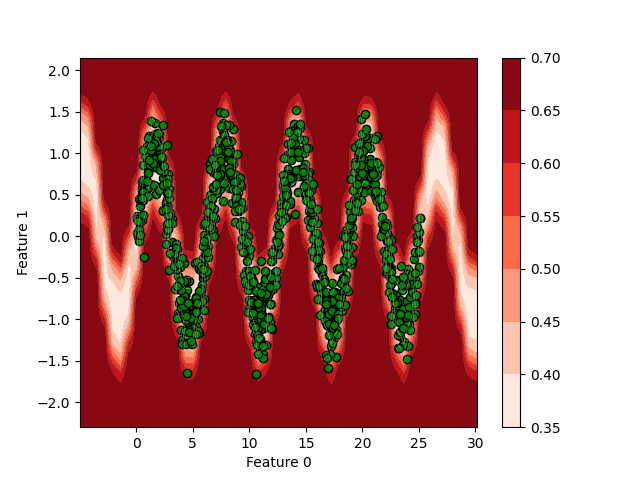}
      \caption{$(x_2-sin(x_1))$-IF}
      \label{fig:hyper-EIFplus}
    \end{subfigure}\hfil % <-- added
\caption{Anomaly scoremaps generated by different IF-based models, where darker colors indicate more anomalous regions.
}\label{fig:AnomalyScoremaps}
\end{figure}

If dataset $X \subset \mathbb{R}^2$ lies on the sine function graph, the optimal $\mathcal{F}$ would contain $f(x_1,x_2) = x_2 - \sin(x_1)$. While this FuBIF effectively finds points deviating from the sine graph, it performs poorly on other structures. Figure~\ref{fig:AnomalyScoremaps} shows how $f(x_1,x_2) = x_2 - \sin(x_1)$ adapts to the point manifold.
This suggests that expanding the set of splitting functions without a balanced probability distribution is not an effective strategy for improving performance and may introduce undesirable biases.

\begin{figure}[t] 
    \centering % <-- added
    \begin{subfigure}{0.45\linewidth}
      \includegraphics[trim={1.5cm 0.7cm 1.5cm 1.2cm}, clip, width=\linewidth]{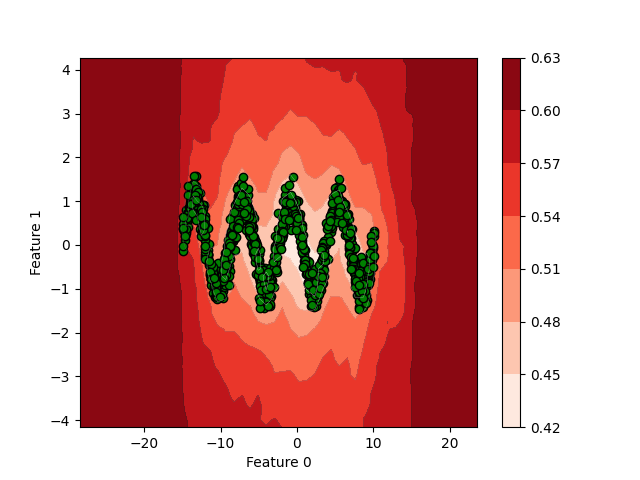}
      \caption{Quadric(1)-IF}
      \label{fig:Quadric(1)-IF}
    \end{subfigure}\hfil % <-- added
    \begin{subfigure}{0.45\linewidth}
      \includegraphics[trim={1.5cm 0.7cm 1.5cm 1.2cm}, clip, width=\linewidth]{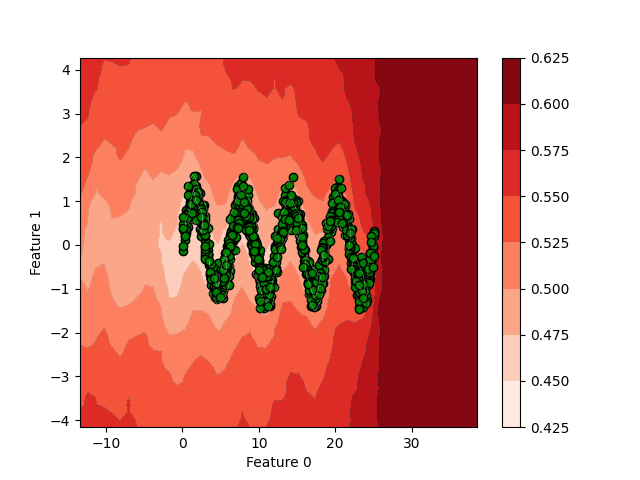}
      \caption{Quadric(1)-IF}
      \label{fig:Quadric(1)-IF-translated}
    \end{subfigure}\hfil % <-- added
    \begin{subfigure}{0.45\linewidth}
      \includegraphics[trim={1.5cm 0.7cm 1.5cm 1.2cm}, clip, width=\linewidth]{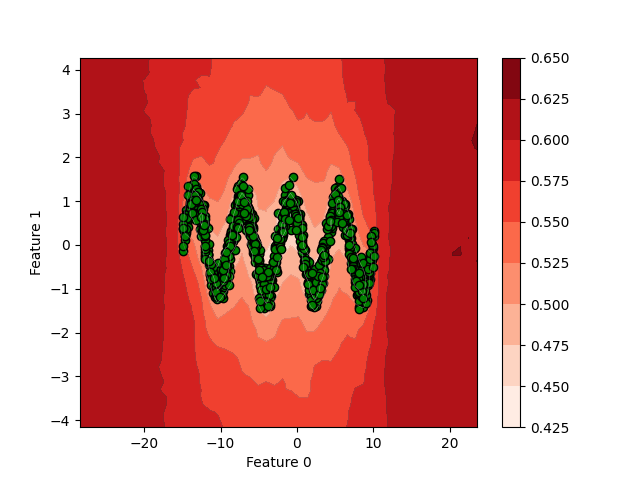}
      \caption{Quadric(100)-IF}
      \label{fig:Quadric(100)-IF}
    \end{subfigure}\hfil % <-- added
    \begin{subfigure}{0.45\linewidth}
      \includegraphics[trim={1.5cm 0.7cm 1.5cm 1.2cm}, clip, width=\linewidth]{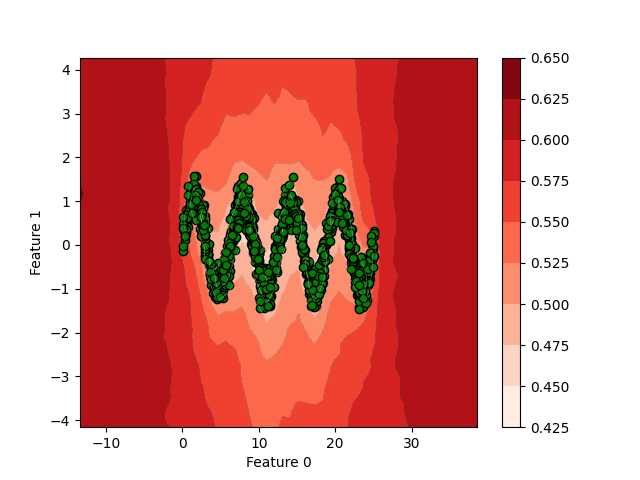}
      \caption{Quadric(100)-IF}
      \label{fig:Quadric(100)-IF-translated}
    \end{subfigure}\hfil % <-- added
    \caption{The images show anomaly scoremaps generated by Quadric($\lambda$)-IF models for $\lambda=1$ and $\lambda=100$. The right column displays the dataset translated left by 10 units with highlighted bias.}\label{fig:bias_analysis}
\end{figure}

Dataset manifold information is often unavailable, requiring $\mathcal{F}$ to cover various distributions without biases. Literature led to the EIF algorithm addressing IF's coordinate axes bias. While defining biases in FuBIF is challenging, AD should remain invariant to transformations such as translations and rotations. 
IF's limitations stem from lack of rotational invariance, while Quadric(1)-IF lacks translational invariance. Figure \ref{fig:bias_analysis} shows that low $\lambda$ values produce lower scores near the axis center.

\subsection{Function-based Isolation Forest Feature Importance.}

FuBIF Feature Importance (FuBIFFI),
inspired by DIFFI \cite{carletti2023interpretable} and ExIFFI \citep{arcudi2023exiffi}, aims to detect the more relevant features for anomaly detection when any  $(\mathcal{F}, \rho, \mu)$-IF is applied.

\begin{figure}[t] 
    \centering % <-- added
    \begin{subfigure}{0.33\linewidth}
      \includegraphics[trim={1.5cm 0.7cm 1.5cm 1.2cm}, clip, width=\linewidth]{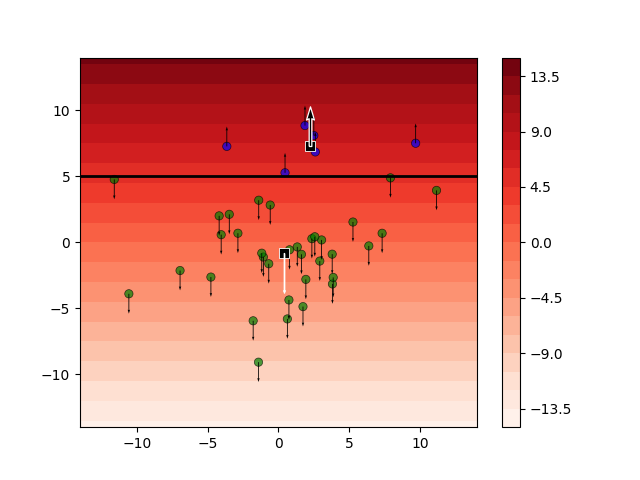}
      \caption{IF}
      \label{fig:hyper-EIF}
    \end{subfigure}\hfil % <-- added
    \begin{subfigure}{0.33\linewidth}
      \includegraphics[trim={1.5cm 0.7cm 1.5cm 1.2cm}, clip, width=\linewidth]{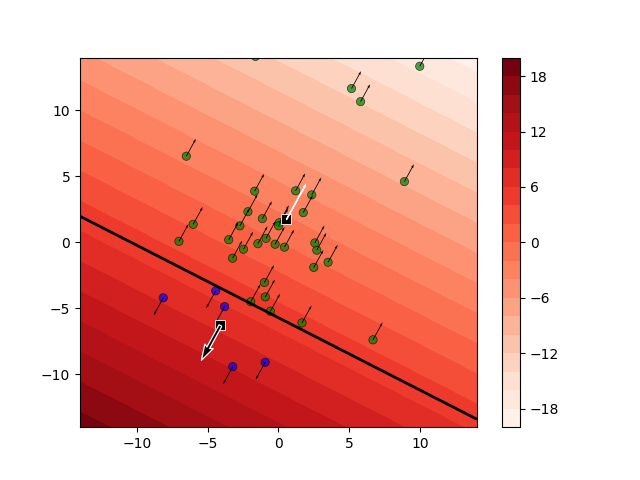}
      \caption{EIF}
      \label{fig:hyper-EIFplus}
    \end{subfigure}\hfil % <-- added
    \begin{subfigure}{0.33\linewidth}
      \includegraphics[trim={1.5cm 0.7cm 1.5cm 1.2cm}, clip, width=\linewidth]{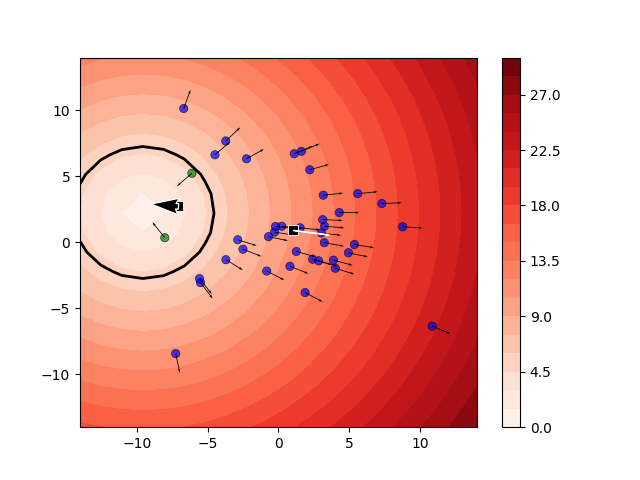}
      \caption{HIF}
      \label{fig:sphereFI}
    \end{subfigure}\hfil % <-- added
    \begin{subfigure}{0.33\linewidth}
      \includegraphics[trim={1.5cm 0.7cm 1.5cm 1.2cm}, clip, width=\linewidth]{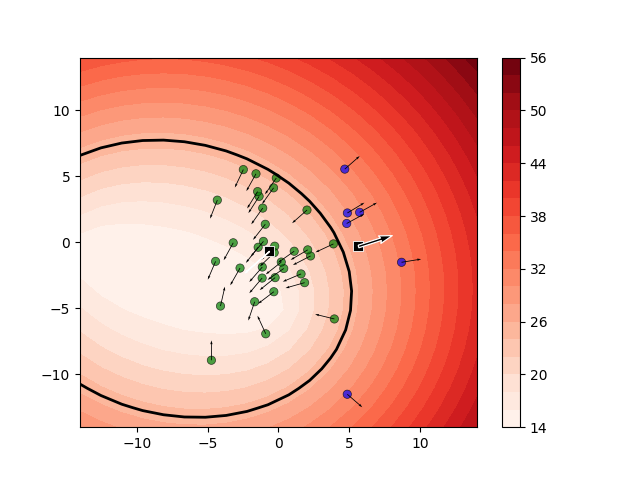}
      \caption{Ellipse$^n$-IF}
      \label{fig:EllipseFI}
    \end{subfigure}\hfil % <-- added
    \begin{subfigure}{0.33\linewidth}
      \includegraphics[trim={1.5cm 0.7cm 1.5cm 1.2cm}, clip, width=\linewidth]{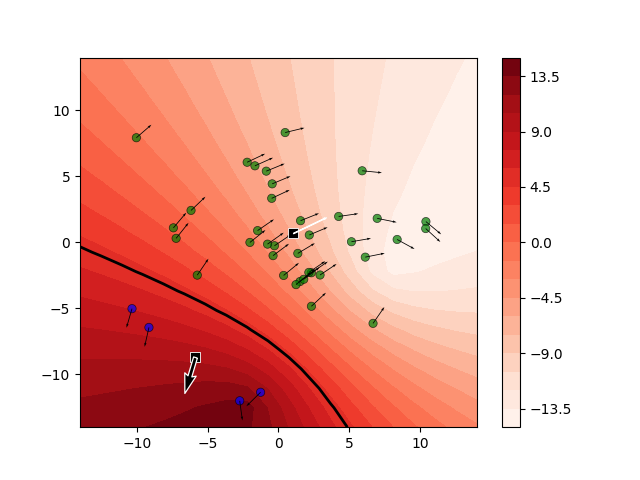}
      \caption{Hyper$^n$-IF}
      \label{fig:hyperbolaFI}
    \end{subfigure}\hfil % <-- added
    \begin{subfigure}{0.33\linewidth}
      \includegraphics[trim={1.5cm 0.7cm 1.5cm 1.2cm}, clip, width=\linewidth]{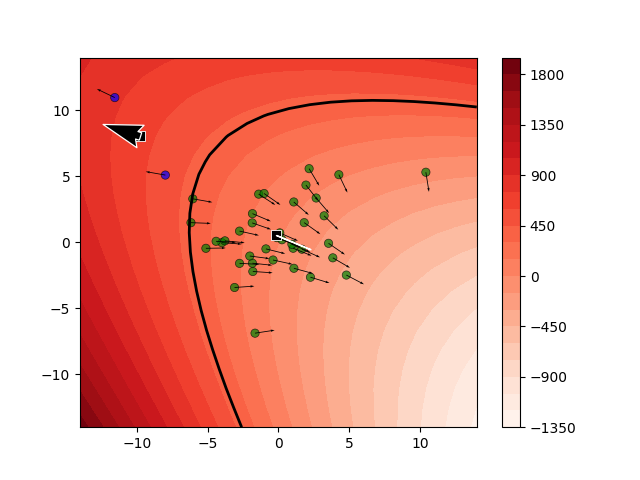}
      \caption{Quad$(\lambda)$-IF}
      \label{fig:quadricFI}
    \end{subfigure}\hfil % <-- added
        \caption{The images show imbalance projections from a branching function toward the mean direction of two subsets. The color map indicates function score, with the black line marking the threshold. Green and blue points represent the subsets and their derivative directions. The thick black arrow shows the mean derivative direction, its thickness reflecting the imbalance in each subset.}\label{fig:f-branching-importances}
\end{figure}

Consider a node $k$ in a tree $T$ where a function $f\in \mathcal{F}$ is chosen. The probability of a point $\mathbf{x}$ being in the right child node depends on the threshold $\tau$ through the distribution $\mu$. This probability is
$$
p_f(\mathbf{x}):=\mathbb{P}_\mu\big(\tau\leq f(\mathbf{x})\big)=\int_{-\infty}^{f(\mathbf{x})} \mu'(t) \mathrm{d}t.
$$
where $\mu'$ is the probability density function of $\mu$.
The gradient of this function expresses the rate of change of its probability lying in the right child node of $k$. It is computed as:
$$
\nabla p_f(\mathbf{x})=\mu'\big(f(\mathbf{x})\big) \Big(\partial_1 f(\mathbf{x}),\  \cdots,\  \partial_nf(\mathbf{x})\Big).
$$
To represent the percentage influence of each feature on the probability $p_f(\mathbf{x})$, we square all entries and divide by the norm squared:
$$
i_k(\mathbf{x})=\frac{1}{\|\nabla f(\mathbf{x})\|^2}  \Big(\partial_1 f(\mathbf{x})^2,\  \cdots,\  \partial_n f(\mathbf{x})^2\Big)
$$
If $f$ is linear, as in IF or EIF, $\nabla f$ is constant and represents the coordinate vector of the selected feature or vector orthogonal to the splitting hyperplane. In these cases, $i_k(\mathbf{x})$ is independent of $\mathbf{x}$. Otherwise, $i_k(\mathbf{x})$ is an $n$-valued function of $\mathbf{x}$.
FuBIFFI uses $i_k(\mathbf{x})$ to produce, for every tree $T$ in the forest, at each node $k$ in the path $\mathcal{P}_T(\mathbf{x})$, a constant vector with positive coordinates summing to $1$, representing the relative importance of each feature in the splitting at node $k$. This vector is defined as:
$$
\mathbf{i}^k_{ \mathbf{x}}=\frac{1}{|Y_{k+1}|}\sum_{\mathbf{y}\in Y_{k+1}} i_k(\mathbf{y})
$$
where $Y_{k+1}$ is the set of points in the next node of $\mathcal{P}_T(\mathbf{x})$.
To evaluate feature importance on the whole path $\mathcal{P}_T(\mathbf{x})$, vectors $\mathbf{i}^k_{\mathbf{x}}$ are summed with weights $\lambda^k_{\mathbf{x}}$, emphasizing nodes that effectively isolate $\mathbf{x}$:
$$
 \lambda^k_{\mathbf{x}}=\frac{|Y_k|}{|Y_{k+1}|+1},
$$
\newpage
Define
$$
\mathbf{i}^T_{\mathbf{x}}= \frac{1}{|\mathcal{P}_T(\mathbf{x})|}\sum_{k\in \mathcal{P}_T(\mathbf{x})} \lambda^k_{\mathbf{x}}\ \mathbf{i}^k_{\mathbf{x}}.
$$
This is the vector of feature importance of $\mathbf{x}$ w.r.t. tree $T$.
Define 
$$
\mathbf{I}_\mathbf{x}=\frac{1}{|\mathcal{T}|}\sum_{T\in\mathcal{T}} \mathbf{i}^T_{\mathbf{x}}.
$$
where $\mathcal{T}$ is the whole forest. This is the vector of local feature importance of $\mathbf{x}$. The importance score of a feature is its corresponding coordinate in $\mathbf{I}_\mathbf{x}$.
To evaluate overall feature importance, the dataset is divided into inliers $I$ and outliers $O$. Define:
$$
\mathbf{I}_I=\frac{1}{|I|}\sum_{\mathbf{x}\in I} \mathbf{I}_\mathbf{x}, \quad \mathbf{I}_O=\frac{1}{|O|}\sum_{\mathbf{x}\in O} \mathbf{I}_\mathbf{x}.
$$
The global feature importance vector is the component-wise ratio of $\mathbf{I}_O$ and $\mathbf{I}_I$:
$$
GFI=\frac{\mathbf{I}_O}{\mathbf{I}_I}
$$
The importance score of a feature is its corresponding coordinate in $GFI$, representing its relevance in separating inliers from outliers.
Figure \ref{fig:f-branching-importances} shows the splitting functions and derivative directions, demonstrating how FuBIFFI works.

\section{Experiments}\label{sec:experiments} 
This section evaluates various FuBIF models from Table \ref{tab:splitting_functions} and established models from {\ttfamily pyod} library \citep{zhao2019pyod}, including IForest \citep{liu2008isolation}, EIF \citep{hariri2019extended}), HIF \citep{choudhury2021hypersphere}, DIF \citep{Xu2022DeepIF}, and AutoEncoder (AE) \citep{aggarwal2016outlier}. These models are tested on benchmark datasets for unsupervised AD, comparing anomaly scores with ground truth.

FuBIFFI reliability is assessed using Feature Selection as a proxy task; and we provide its time complexity analysis.
We also examine NN-IF, whose splitting function is a randomly initialized NN (see Table \ref{tab:splitting_functions}).

\subsection{Dataset} 
We use AD tabular datasets from the ``ODDS" library \citep{Rayana:2016}, which contains real-world AD datasets derived from multi-class datasets where the least represented class becomes an outlier, with others merged as inlier class. The datasets appear in Table \ref{tab:dataset_used}. 

Synthetic datasets {\ttfamily Xaxis} and {\ttfamily Bisect\_3d} are also used. Both are based on a 6-dimensional hypersphere centered at zero, with anomalies along the first feature and the first three features respectively, demonstrating bias in certain splitting functions like IF. 
\begin{table}[htbp]
\centering
\caption{Experimental datasets overview.}
\label{tab:dataset_used}
\footnotesize
\resizebox{\linewidth}{!}{%
\begin{tabular}{|c|cccccc|}
% \toprule
\hline
{} &  data  &  anomalies & contam  &  features & Size &  Type \\
{} &  $n$  &    & \%  &  $d$ & &\\
\hline
\texttt{Xaxis} & 1100 & 100 & 9.09 & 6 & (Low) & Synt \\
\texttt{Bisec3D} & 1100 & 100 &  9.09 & 6 & (Low) & Synt \\
\texttt{Annthyroid} & 7200 & 534 & 7.56 & 6 & (Low) & Real \\
\texttt{Breastw} & 683 & 239  & 52.56 & 9 & (Middle) & Real \\
\texttt{Cardio} & 1831 & 176 & 9.60 & 21 & (High) & Real \\
\texttt{Glass} & 214& 29 & 13.55& 9& (Middle) & Real \\
\texttt{Ionosphere} & 351 & 126  &  35.71 & 33 & (High) & Real\\
\texttt{Pendigits} & 6870 & 156 & 2.27 & 16  & (Middle) & Real\\
\texttt{Pima} & 768 & 268 & 34.89 & 8 & (Middle) & Real\\
\texttt{Wine} & 129	&  10 &  7.75	& 13   & (Middle) & Real \\ 
\hline
\end{tabular}
}
\end{table}

% ========================================================================================

\subsection{Methodology}
Inspired by \cite{pang2019deep,Xu2022DeepIF}, we observed that model performance varies with contamination levels in training data. For real datasets, higher contamination causes IF-based models to include anomalies into the inlier distribution, misclassifying peripheral points as outliers.

Models were evaluated in two training scenarios: \textbf{Scenario I}, where all data, including anomalies, were used; and \textbf{Scenario II}, where only inliers were used for training.

Two threshold strategies for the splitting function (Section~\ref{mathframe}) were also tested. The \(\mathcal{U}\) distribution restricts cuts within the data range; \(\mathcal{N}\) permits extrapolated cuts, potentially forming empty branches and improving the detection of unseen anomalies.

Performance was measured via average precision (Avg Prec) and ROC AUC, independent of contamination. Precision based on known contamination was also reported.
% reference to tab:avg-prec-total which was in the appendix
% Since IF-based models are stochastic, all results in Table \ref{tab:avg-prec-total} and Figure \ref{fig:interpretation} are averages of 10 runs.
Since IF-based models are stochastic, in Figure \ref{fig:interpretation} we show an average of 10 runs.

Avg Precision evaluates positive prediction quality by combining precision and recall across thresholds. ROC AUC measures class discrimination ability across thresholds, showing the difference between precise anomaly detection and overall discrimination.

To evaluate the interpretation algorithms, we used the $AUC_{FS}$ metric by \cite{turbe2023evaluation} and refined by \cite{arcudi2023exiffi} for Feature Importance in EIF. This metric measures the area between feature selection processes by comparing the effect of dropping the least versus most important features, and assesses the robustness and reliability in evaluating feature importance.

\subsection{Results}
\subsubsection{Performance Evaluation}
Table \ref{tab:small_table} shows the performance metrics of the models from Section \ref{sec:experiments} on the datasets in Scenario II, where the threshold \( \tau \) is selected using the normal distribution \(\mathcal{N}\). 
% Complete results can be found in Table \ref{tab:avg-prec-total} in Appendix \ref{app:complete_results}.
% 
\begin{table}[ht]
\centering
\caption{Avg precision results of the models in Scenario 2 with Normal threshold distribution}
\label{tab:small_table}
\resizebox{\columnwidth}{!}{%
\begin{tabular}{|l|>{\arraybackslash}p{0.65cm}|>{\arraybackslash}p{0.65cm}|>{\arraybackslash}p{0.65cm}|>{\arraybackslash}p{0.65cm}|>{\arraybackslash}p{0.65cm}|>{\arraybackslash}p{0.65cm}|>{\arraybackslash}p{0.65cm}|>{\arraybackslash}p{0.65cm}|>{\arraybackslash}p{0.65cm}|>{\arraybackslash}p{0.65cm}|}
\hline
% -- original
% \textbf{Dataset} & \rotatebox{90}{\textbf{IF}} & \rotatebox{90}{\textbf{EIF}} & \rotatebox{90}{\textbf{HIF}} & \rotatebox{90}{\textbf{Ellipsoid-IF}} & \rotatebox{90}{\textbf{Paraboloid-IF}} & \rotatebox{90}{\textbf{Hyperboloid-IF}} & \rotatebox{90}{\textbf{Quadric-IF}} & \rotatebox{90}{\textbf{NN-IF}} & \rotatebox{90}{\textbf{DIF}} & \rotatebox{90}{\textbf{AutoEncoder}} \\
% -- short consistent
% \textbf{Dataset} & \rotatebox{90}{\textbf{IF}} & \rotatebox{90}{\textbf{EIF}} & \rotatebox{90}{\textbf{HIF}} & \rotatebox{90}{\textbf{Ellipse$^n$-IF}} & \rotatebox{90}{\textbf{Para$^n$-IF}} & \rotatebox{90}{\textbf{Hyper$^n$-IF}} & \rotatebox{90}{\textbf{Quad$(\lambda)$-IF}} & \rotatebox{90}{\textbf{NN-IF}} & \rotatebox{90}{\textbf{DIF}} & \rotatebox{90}{\textbf{AE}} \\
% shorter but slightly inconsistent naming
% \textbf{Dataset} & \rotatebox{90}{\textbf{IF}} & \rotatebox{90}{\textbf{EIF}} & \rotatebox{90}{\textbf{HIF}} & \rotatebox{90}{\textbf{Ellipse-IF}} & \rotatebox{90}{\textbf{Para-IF}} & \rotatebox{90}{\textbf{Hyper-IF}} & \rotatebox{90}{\textbf{Quad-IF}} & \rotatebox{90}{\textbf{NN-IF}} & \rotatebox{90}{\textbf{DIF}} & \rotatebox{90}{\textbf{AE}} \\
% shorter but slightly inconsistent naming - horizontally centered
\textbf{Dataset} & \makebox[0.65cm][c]{\rotatebox{90}{\textbf{IF}}} & \makebox[0.65cm][c]{\rotatebox{90}{\textbf{EIF}}} & \makebox[0.65cm][c]{\rotatebox{90}{\textbf{HIF}}} & \makebox[0.65cm][c]{\rotatebox{90}{\textbf{Ellipse-IF}}} & \makebox[0.65cm][c]{\rotatebox{90}{\textbf{Para-IF}}} & \makebox[0.65cm][c]{\rotatebox{90}{\textbf{Hyper-IF}}} & \makebox[0.65cm][c]{\rotatebox{90}{\textbf{Quad-IF}}} & \makebox[0.65cm][c]{\rotatebox{90}{\textbf{NN-IF}}} & \makebox[0.65cm][c]{\rotatebox{90}{\textbf{DIF}}} & \makebox[0.65cm][c]{\rotatebox{90}{\textbf{AE}}} \\
\hline
\texttt{Xaxis}       & 0.06 & \textbf{1.00} & \textbf{1.00} & \textbf{1.00} & \textbf{1.00} & \textbf{1.00} & \textbf{1.00} & \textbf{1.00} & \textbf{1.00} & \textbf{1.00} \\
\texttt{bisect\_3d}  & \textbf{1.00} & \textbf{1.00} & \textbf{1.00} & \textbf{1.00} & \textbf{1.00} & \textbf{1.00} & \textbf{1.00} & \textbf{1.00} & \textbf{1.00} & \textbf{1.00} \\
\texttt{annthyroid}  & 0.47 & 0.48 & \textbf{0.52} & \textbf{0.52} & 0.50 & 0.44 & 0.48 & 0.49 & 0.31 & 0.44 \\
\texttt{breastw}     & \textbf{0.99} & \textbf{0.99} & \textbf{0.99} & \textbf{0.99} & \textbf{0.99} & \textbf{0.99} & \textbf{0.99} & 0.96 & 0.94 & \textbf{0.99} \\
\texttt{cardio}      & 0.67 & 0.77 & 0.76 & 0.77 & 0.77 & 0.64 & 0.77 & 0.32 & \textbf{0.79} & 0.76 \\
\texttt{glass}       & 0.82 & 0.77 & 0.88 & \textbf{0.90} & 0.85 & 0.71 & 0.68 & 0.67 & 0.71 & 0.58 \\
\texttt{ionosphere}  & 0.92 & \textbf{0.97} & 0.96 & 0.95 & \textbf{0.97} & 0.96 & 0.96 & 0.94 & 0.93 & 0.87 \\
\texttt{pendigits}   & \textbf{0.47} & 0.39 & 0.32 & 0.29 & 0.37 & 0.25 & 0.37 & 0.02 & 0.41 & 0.24 \\
\texttt{pima}        & \textbf{0.64} & 0.60 & 0.61 & 0.62 & 0.61 & 0.57 & 0.59 & 0.55 & 0.56 & 0.52 \\
\texttt{wine}        & 0.76 & 0.85 & 0.81 & 0.81 & \textbf{0.86} & 0.69 & \textbf{0.86} & 0.67 & 0.82 & 0.48 \\
\hline
\textbf{Mean}        & 0.68 & 0.78 & 0.79 & 0.79 & 0.79 & 0.73 & 0.79 & 0.71 & 0.76 & 0.73 \\
\hline
\end{tabular}
}
\end{table}
The average performance across all IF-based models is similar, with effectiveness varying by dataset.
Based on results we provide in our open-source repository, FuBIF algorithms outperform both AE and DIF models, except for the {\ttfamily glass} and {\ttfamily cardio} datasets, even if no branching function consistently outperformed the rest. IF is 1.5 times more effective than Ellipse-IF on the {\ttfamily pendigits} dataset because of its square-shaped distribution. In contrast, Ellipse-IF performs better in the {\ttfamily wine} and {\ttfamily Xaxis} datasets, where the data have a globular distribution. This suggests that branching function selection should be dataset-specific for optimal results.

\subsubsection{Interpretability Evaluation}
\begin{figure}[htbp] 
    \centering % <-- added
      \includegraphics[width=0.7\linewidth]{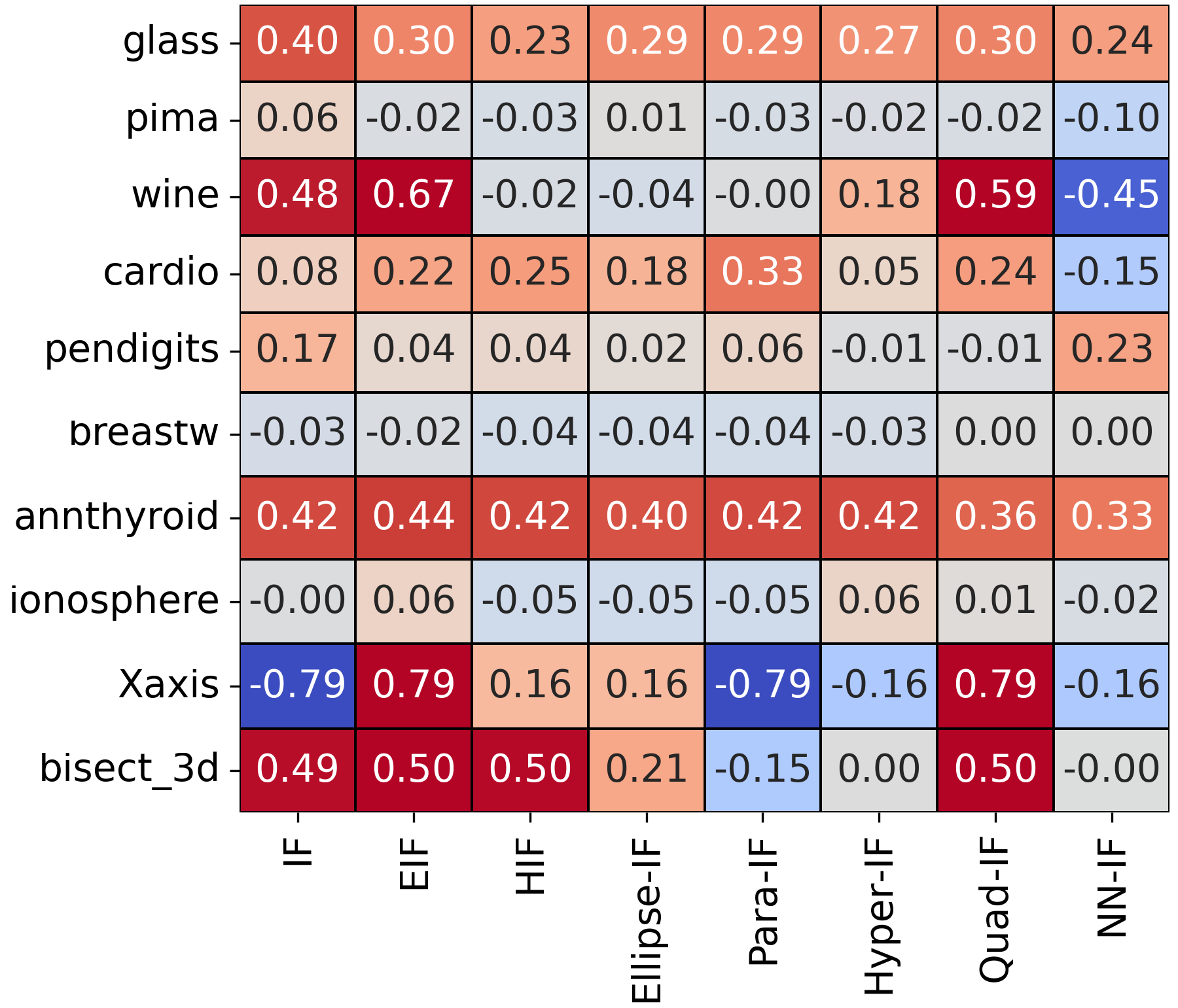}
    \caption{Heatmap of $AUC_{FS}$ scores for datasets in Scenario II with normal threshold distribution. Higher score indicates better feature selection.}\label{fig:interpretation}
\end{figure}

The evaluation of the interpretation algorithm shown in Figure \ref{fig:interpretation} demonstrates, through the feature selection performance ($AUC_{FS}$ score), that it remains consistent across models, with some notable exceptions, indicating a need for refinement.
The models show varying efficiencies in determining feature importance across datasets, with notable differences between {\ttfamily glass} and {\ttfamily wine}.

The results show that although no branching function is universally superior, the interpretation algorithm consistently performs feature selection across models and datasets. This robustness demonstrates its effectiveness in identifying the key features and their importance. This insight helps guide branching function selection, align the model structure with domain knowledge, and ensure logical and practical relevance.

\subsection{Computational Time Analysis}
\begin{figure}[ht!]
    \centering % <-- added
    \begin{subfigure}{0.7\linewidth}
      \includegraphics[trim={0.0cm 0.0cm 0.0cm 0.0cm}, clip, width=\linewidth]{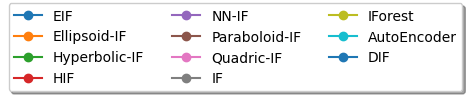}
      \caption{Legend}
      \label{fig:legend}
    \end{subfigure}\hfil % <-- added
    \centering % <-- added
    \begin{subfigure}{0.5\linewidth}
      \includegraphics[trim={0.0cm 0.0cm 0.0cm 2.6cm}, clip, width=\linewidth]{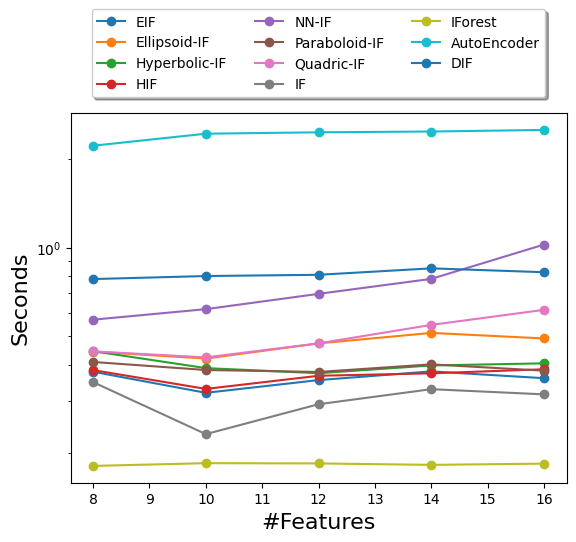}
      \caption{Fit VS Features}
      \label{fig:fit_feat}
    \end{subfigure}\hfil % <-- added
    \begin{subfigure}{0.5\linewidth}
      \includegraphics[trim={0.0cm 0.0cm 0.0cm 2.6cm}, clip, width=\linewidth]{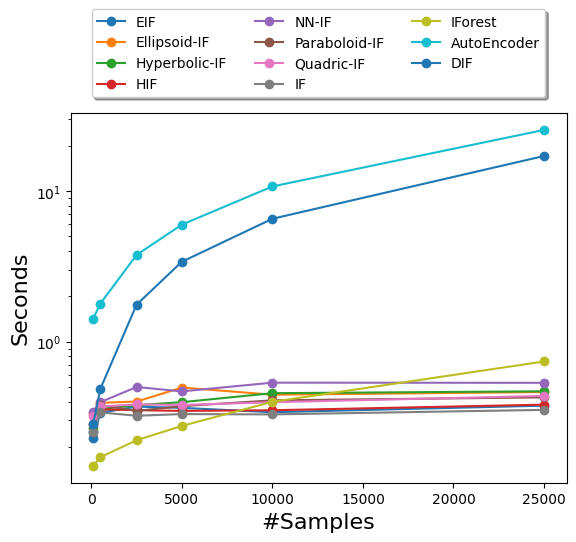}
      \caption{Fit VS Samples}
      \label{fig:fit_samp}
    \end{subfigure}\hfil % <-- added
    \begin{subfigure}{0.5\linewidth}
      \includegraphics[trim={0.0cm 0.0cm 0.0cm 2.6cm}, clip, width=\linewidth]{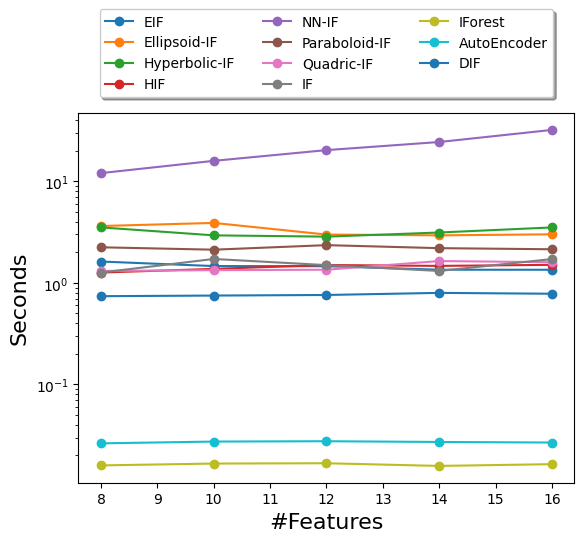}
      \caption{Predict VS Features}
      \label{fig:hyper-EIF}
    \end{subfigure}\hfil % <-- added
    \begin{subfigure}{0.5\linewidth}
      \includegraphics[trim={0.0cm 0.0cm 0.0cm 2.6cm}, clip, width=\linewidth]{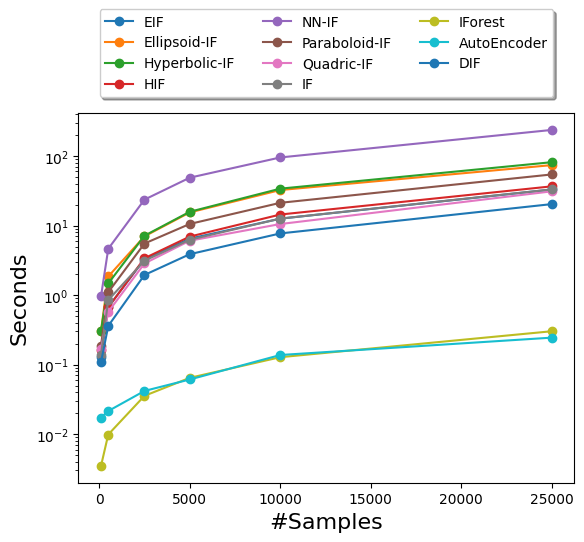}
      \caption{Predict VS Samples}
      \label{fig:hyper-EIFplus}
    \end{subfigure}\hfil % <-- added
    \begin{subfigure}{0.5\linewidth}
      \includegraphics[trim={0.0cm 0.0cm 0.0cm 2.6cm}, clip, width=\linewidth]{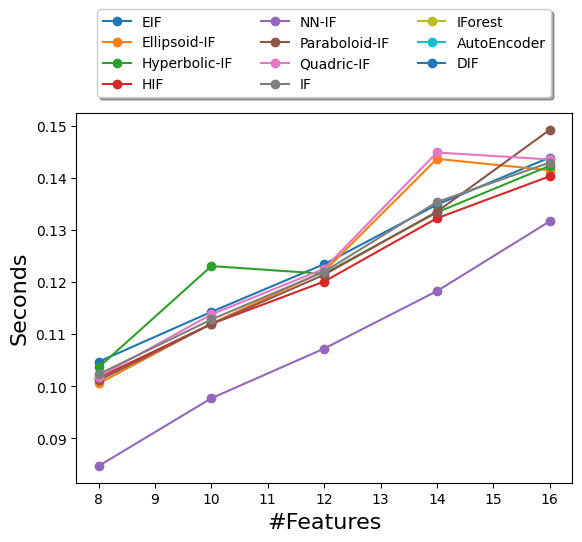}
      \caption{Interpretation VS Features}
      \label{fig:hyper-EIFplus}
    \end{subfigure}\hfil % <-- added
    \begin{subfigure}{0.5\linewidth}
      \includegraphics[trim={0.0cm 0.0cm 0.0cm 2.6cm}, clip, width=\linewidth]{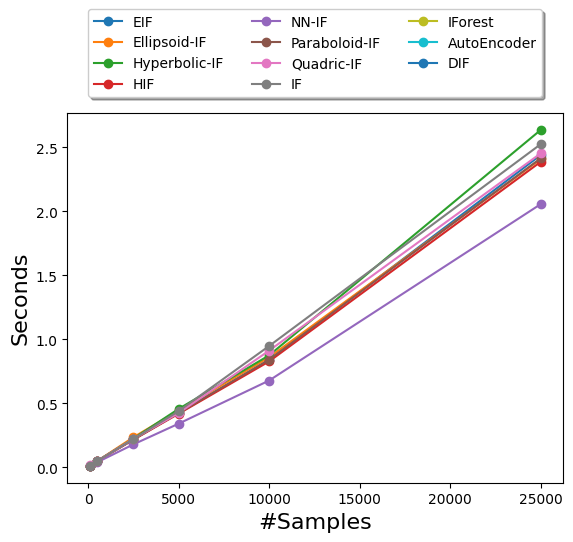}
      \caption{Interpretation VS Samples}
      \label{fig:hyper-EIFplus}
    \end{subfigure}\hfil % <-- added
        \caption{Computational time of the models varying the size and dimensionality of the dataset.}\label{fig:time_analysis}
\end{figure}

In Figure \ref{fig:time_analysis} we analyze the time complexity of the FuBIF models compared to other algorithms. All FuBIF algorithms show similar time complexity, with slight increases when more parameters are required for branching. FuBIF models' fitting time scales better than AutoEncoder and DIF, although prediction time exceeds DIF. Function-based branching maintains an efficient computation and minimal memory requirements.

\subsection{Neural Network Branching}
Inspired by \cite{Xu2022DeepIF}, we use a randomly initialized Neural Network (NN) as each node's branching function. Unlike DIF, our approach assigns a unique NN per node that maps the dataset to a one-dimensional distribution to generate a branching threshold.

\begin{figure}
    \centering % <-- added
    \begin{subfigure}{0.45\linewidth}
      \includegraphics[trim={1.5cm 0.7cm 1.5cm 1.2cm}, clip, width=\linewidth]{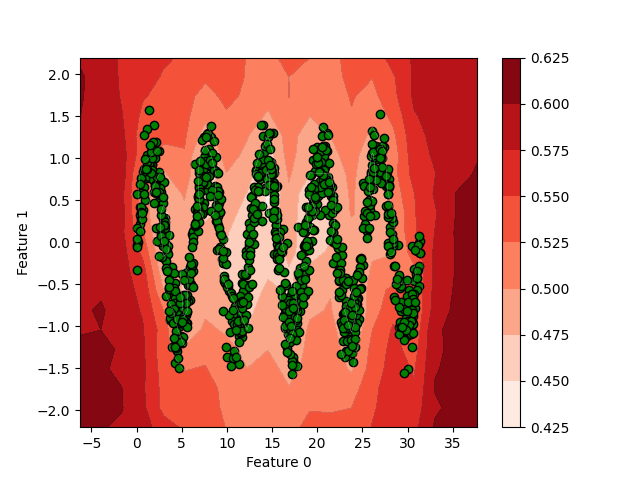}
      \caption{Anomaly Scormap of a NN-IF model}\label{fig:NN-Splitting-Scoremap}
    \end{subfigure}\hfil % <-- added
    \begin{subfigure}{0.45\linewidth}
      \includegraphics[trim={1.5cm 0.7cm 1.5cm 1.2cm}, clip, width=\linewidth]{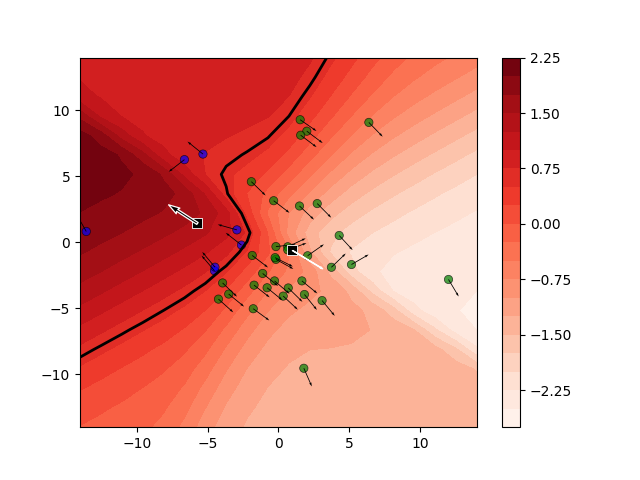}
      \caption{Importances value and direction of a NN branch}\label{fig:NN-branch-importance}
    \end{subfigure}\hfil % <-- added
    \caption{Analysis of NN branching function}\label{fig:NN_analysis}
\end{figure}
The adaptability of the model is evident in Figure \ref{fig:NN-Splitting-Scoremap}, where the importance directions of the cuts in Figure \ref{fig:NN-branch-importance} are aligned with the dataset distribution and branching function. However, because of the complexity of neural networks and chaotic gradients, the results are less interpretable than those of other models, as shown by the lower $AUC_{FS}$ scores in Figure \ref{fig:interpretation}.

\section{Future Works}\label{sec:future_works}
% 
% -- Original Version
% FuBIF introduces a unifying framework that not only opens new avenues for isolation-based AD algorithms but also presents intriguing mathematical challenges. We highlight the following open research directions:

% \textbf{Bias in Algorithms:} Figure \ref{fig:bias_analysis} shows how Quadric(1)-IF exhibited bias, interpreted as a lack of translational invariance, while the lack of rotational invariance in IF led to EIF. Future work should focus on formally defining these biases and developing symmetrization techniques to mitigate them.
% \\
% \textbf{Adaptive FuBIF:} Efforts should aim at better methods for selecting \(f\) from \(\mathcal{F}\). One approach is adapting \(\mathcal{F}\) to randomly chosen subsets of the data, selecting functions that best fit the dataset.
% \\
% \textbf{Ensemble Methods:} Using different branching functions in a FuBIF ensemble could enhance insights into the dataset's manifold, enabling effective Manifold Mining.
% 
% -- Reduced version
FuBIF unified framework for isolation-based AD algorithms offers improved capabilities while introducing some mathematical complexities. 
Key research directions include the following.
\\
\textbf{Bias in Algorithms:} Figure \ref{fig:bias_analysis} shows Quadric(1)-IF's lack of translational invariance, while IF's lack of rotational invariance led to EIF. Research should define biases and develop symmetrization techniques.
\\
\textbf{Adaptive FuBIF:} Future work should develop methods for selecting \(f\) from \(\mathcal{F}\), including adapting \(\mathcal{F}\) to data subsets.
\\
\textbf{Ensemble Methods:} Using varied branching functions in FuBIF ensembles could improve dataset manifold understanding.

\section{Conclusion}\label{sec:conclusion}
FuBIF is a unified framework for isolation-based AD, which is generalized and flexible. 
Its mathematical core and adaptability make it a powerful tool for unsupervised anomaly detection, handling complex data, and offering valuable interpretations. 
With diverse branching and thresholding functions, FuBIF addresses limitations in traditional Isolation Forest models with promising results; although its effectiveness varies with data characteristics, showing no universally superior branching function.
FuBIFFI embeds interpretability by providing a generalized derivation of the feature importance scores for isolation-based algorithms. Research directions include mitigating bias, developing distribution-specific models, and evaluating branching functions.

\bibliography{ifacconf}

\begin{thebibliography}{18}
\providecommand{\natexlab}[1]{#1}
\providecommand{\url}[1]{\texttt{#1}}
\providecommand{\urlprefix}{URL }
\expandafter\ifx\csname urlstyle\endcsname\relax
  \providecommand{\doi}[1]{doi:\discretionary{}{}{}#1}\else
  \providecommand{\doi}{doi:\discretionary{}{}{}\begingroup \urlstyle{rm}\Url}\fi

\bibitem[{Aggarwal(2016)}]{aggarwal2016outlier}
Aggarwal, C.C. (2016).
\newblock Outlier analysis second edition.

\bibitem[{Arcudi et~al.(2023)Arcudi, Frizzo, Masiero, and Susto}]{arcudi2023exiffi}
Arcudi, A., Frizzo, D., Masiero, C., and Susto, G.A. (2023).
\newblock Exiffi and eif+: Interpretability and enhanced generalizability to extend the extended isolation forest.
\newblock \emph{arXiv preprint arXiv:2310.05468}.

\bibitem[{Arrieta et~al.(2020)Arrieta, D{\'\i}az-Rodr{\'\i}guez, Del~Ser, Bennetot, Tabik et~al.}]{arrieta2020explainable}
Arrieta, A.B., D{\'\i}az-Rodr{\'\i}guez, N., Del~Ser, J., Bennetot, A., Tabik, S., et~al. (2020).
\newblock Explainable artificial intelligence (xai): Concepts, taxonomies, opportunities and challenges toward responsible ai.
\newblock \emph{Information fusion}, 58, 82--115.

\bibitem[{Carletti et~al.(2023)Carletti, Terzi, and Susto}]{carletti2023interpretable}
Carletti, M., Terzi, M., and Susto, G.A. (2023).
\newblock Interpretable anomaly detection with diffi: Depth-based feature importance of isolation forest.
\newblock \emph{Eng. Applications of A.I.}, 119, 105730.

\bibitem[{Choudhury et~al.(2021)Choudhury, Ky, Ren, and Shi}]{choudhury2021hypersphere}
Choudhury, J., Ky, P., Ren, Y., and Shi, C. (2021).
\newblock Hypersphere for branching node for the family of isolation forest algorithms.
\newblock In \emph{2021 IEEE Int. Conf. on Smart Computing (SMARTCOMP)}, 418--423. IEEE.

\bibitem[{Doshi-Velez and Kim(2017)}]{doshi2017towards}
Doshi-Velez, F. and Kim, B. (2017).
\newblock Towards a rigorous science of interpretable machine learning.
\newblock \emph{arXiv preprint arXiv:1702.08608}.

\bibitem[{Hariri et~al.(2019)Hariri, Kind, and Brunner}]{hariri2019extended}
Hariri, S., Kind, M.C., and Brunner, R.J. (2019).
\newblock Extended isolation forest.
\newblock \emph{IEEE Tran. on knowledge and data Eng.}, 33(4), 1479--1489.

\bibitem[{Kabir et~al.(2023)Kabir, Shufian, and Zishan}]{kabir2023isolation}
Kabir, S., Shufian, A., and Zishan, M.S.R. (2023).
\newblock Isolation forest based anomaly detection and fault localization for solar pv system.
\newblock In \emph{2023 3rd Int. Con. on Robotics, Electrical and Signal Processing Techniques (ICREST)}, 341--345. IEEE.

\bibitem[{Lesouple et~al.(2021)Lesouple, Baudoin, Spigai, and Tourneret}]{lesouple2021generalized}
Lesouple, J., Baudoin, C., Spigai, M., and Tourneret, J.Y. (2021).
\newblock Generalized isolation forest for anomaly detection.
\newblock \emph{Pattern Recognition Letters}, 149, 109--119.

\bibitem[{Liu et~al.(2008)Liu, Ting, and Zhou}]{liu2008isolation}
Liu, F.T., Ting, K.M., and Zhou, Z.H. (2008).
\newblock Isolation forest.
\newblock In \emph{2008 eighth IEEE Int. Conf. on data mining}, 413--422. IEEE.

\bibitem[{Ounacer et~al.(2018)Ounacer, El~Bour, Oubrahim, Ghoumari, and Azzouazi}]{ounacer2018using}
Ounacer, S., El~Bour, H.A., Oubrahim, Y., Ghoumari, M.Y., and Azzouazi, M. (2018).
\newblock Using isolation forest in anomaly detection: the case of credit card transactions.
\newblock \emph{Periodicals of Eng. and Natural Sciences}, 6(2), 394--400.

\bibitem[{Pang et~al.(2019)Pang, Shen, and Van Den~Hengel}]{pang2019deep}
Pang, G., Shen, C., and Van Den~Hengel, A. (2019).
\newblock Deep anomaly detection with deviation networks.
\newblock In \emph{Proceedings of the 25th ACM SIGKDD international conference on knowledge discovery \& data mining}, 353--362.

\bibitem[{Rayana(2016)}]{Rayana:2016}
Rayana, S. (2016).
\newblock {ODDS} library.
\newblock \urlprefix\url{https://odds.cs.stonybrook.edu}.

\bibitem[{Rudin(2019)}]{rudin2019stop}
Rudin, C. (2019).
\newblock Stop explaining black box machine learning models for high stakes decisions and use interpretable models instead.
\newblock \emph{Nature machine intelligence}, 1(5), 206--215.

\bibitem[{Turb{\'e} et~al.(2023)Turb{\'e}, Bjelogrlic, Lovis, and Mengaldo}]{turbe2023evaluation}
Turb{\'e}, H., Bjelogrlic, M., Lovis, C., and Mengaldo, G. (2023).
\newblock Evaluation of post-hoc interpretability methods in time-series classification.
\newblock \emph{Nature Machine Intelligence}, 5(3), 250--260.

\bibitem[{Wu and Chen(2018)}]{wu2018application}
Wu, W. and Chen, Y. (2018).
\newblock Application of isolation forest to extract multivariate anomalies from geochemical exploration data.
\newblock \emph{Global Geology}, 21(1), 36--47.

\bibitem[{Xu et~al.(2022)Xu, Pang, Wang, and Wang}]{Xu2022DeepIF}
Xu, H., Pang, G., Wang, Y., and Wang, Y. (2022).
\newblock Deep isolation forest for anomaly detection.
\newblock \emph{IEEE Tran. on Knowledge and Data Eng.}, 35, 12591--12604.

\bibitem[{Zhao et~al.(2019)Zhao, Nasrullah, and Li}]{zhao2019pyod}
Zhao, Y., Nasrullah, Z., and Li, Z. (2019).
\newblock Pyod: A python toolbox for scalable outlier detection.
\newblock \emph{Journal of Machine Learning Research}, 20(96), 1--7.
\newblock \urlprefix\url{http://jmlr.org/papers/v20/19-011.html}.

\end{thebibliography}
\end{document}